\documentclass{article}

\usepackage[utf8]{inputenc}
\usepackage{amsfonts}
\usepackage{nicefrac}
\usepackage{microtype} 
\usepackage{float, array}
\usepackage{amssymb, amsmath, bm, amsthm}
\usepackage[caption=false,font=scriptsize]{subfig}
\usepackage[ruled,vlined]{algorithm2e}
\usepackage{enumerate}
\usepackage{color, colortbl}
\usepackage{booktabs, multirow}
\definecolor{darkgreen}{rgb}{0, 0.5, 0}
\definecolor{red}{rgb}{1, 0, 0}
\definecolor{purple}{rgb}{0.5, 0, 0.5}
\usepackage{chngpage}

\usepackage{mfirstuc}
\usepackage{mathtools}
\usepackage{lettrine}
\usepackage{pxfonts}

\newcommand\ie{\textit{i.e.}}
\newcommand\eg{\textit{e.g.}}

\usepackage[square,sort,comma,numbers]{natbib}
\usepackage[preprint]{neurips_2019}
\usepackage{amsmath,amsthm,amssymb}
\usepackage{physics}

\usepackage[utf8]{inputenc} % allow utf-8 input
\usepackage[T1]{fontenc}    % use 8-bit T1 fonts
\usepackage{url}            % simple URL typesetting
\usepackage{booktabs}       % professional-quality tables
\usepackage{amsfonts}       % blackboard math symbols
\usepackage{nicefrac}       % compact symbols for 1/2, etc.
\usepackage{microtype}      % microtypography
\usepackage{comment}

\title{Revisit Policy Optimization in Matrix Form}

\author{
	Sitao Luan$^{1,2}$, Xiao-Wen Chang$^{1}$, Doina Precup$^{1,2,3}$\\
	\{sitao.luan@mail, chang@cs, dprecup@cs\}.mcgill.ca\\
	$^1$McGill University; $^2$Mila; $^3$DeepMind
}

\begin{document}
	
	\maketitle
	
	\begin{abstract}
		In tabular case, when the reward and environment dynamics are known, policy evaluation can be written as $\bm{V}_{\bm{\pi}} = (I - \gamma P_{\bm{\pi}})^{-1} \bm{r}_{\bm{\pi}}$, where $P_{\bm{\pi}}$ is the state transition matrix given policy ${\bm{\pi}}$ and $\bm{r}_{\bm{\pi}}$ is the reward signal given ${\bm{\pi}}$. What annoys us is that $P_{\bm{\pi}}$ and $\bm{r}_{\bm{\pi}}$ are both mixed with ${\bm{\pi}}$, which means every time when we update ${\bm{\pi}}$, they will change together. In this paper, we leverage the notation from \cite{wang2007dual} to disentangle ${\bm{\pi}}$ and environment dynamics which makes optimization over policy more straightforward. We show that policy gradient theorem \cite{sutton2018reinforcement} and TRPO \cite{schulman2015trust} can be put into a more general framework and such notation has good potential to be extended to model-based reinforcement learning.
	\end{abstract}
	\section{Introduction}
	\subsection{Bellman Equation in Matrix Form}
	
	Markov decision process (MDP) is a framework to model the learning process that the agent learns from the interaction with the environment \cite{sutton2018reinforcement}. The interaction happens in discrete time steps, $t=0,1,2,3,\cdots$. At step $t$, given a state $S_t = s_t \in \mathcal{S}$, the agent picks an action $a_t \in \mathcal{A}(s_t)$ according to a policy ${\bm{\pi}}(\cdot | s_t)$, which is a rule of choosing actions given a state. Then, at time $t+1$, the environmental dynamics $p: \mathcal{S} \times \mathcal{R} \times \mathcal{A} \times \mathcal{S} \rightarrow [0,1]$ takes the agent to a new state $S_{t+1} = s_{t+1} \in \mathcal{S}$ and provide a numerical reward $R_{t+1} = r_{t+1}(s_t, a_t, s_{t+1}) \in  \mathbb{R}$. Such a sequence of interactions gives us a trajectory $\tau = \{S_0, A_0,R_1,S_1,A_1,R_2,S_2,A_2,R_3,\cdots\}$. Our objective is to find an optimal policy to maximize the expected long-term discounted cumulative rewards $\bm{V}_{\bm{\pi}}(s) = E_{\bm{\pi}}[\sum_{k=0}^{\infty} \gamma^k R_{t+k+1}|S_t=s]$ for each state $s$ or $\bm{Q}_{\bm{\pi}}(s,a) = E_{\bm{\pi}} [\sum_{k=0}^{\infty} \gamma^k R_{t+k+1}|S_t = s,a_t = a]$ for each state-action pair $(s,a)$, where $\gamma$ is the discount factor. 
	The Bellman equation for $\bm{V}_{\bm{\pi}}(s)$ can be written as follows:
	\begin{equation}
	\begin{aligned} \label{elementwise bellman equation}
	\bm{V}_{\bm{\pi}}(s)&=\sum_a {\bm{\pi}}(a|s) \sum_{s',r}  p(s',r|s,a)[r+\gamma \bm{V}_{\bm{\pi}}(s')] \\ 
	&=\sum_a {\bm{\pi}}(a|s) \sum_{s',r}  r\cdot p(s',r|s,a)+\sum_a {\bm{\pi}}(a|s) \sum_{s',r} p(s',r|s,a) \gamma \bm{V}_{\bm{\pi}}(s') \\ 
	&=\sum_a {\bm{\pi}}(a|s) r(s,a) + \gamma  \sum_{s'} \{\sum_a  {\bm{\pi}}(a|s) \cdot p(s'|s,a) \} \bm{V}_{\bm{\pi}}(s')\\
	&=\bm{r}_{\bm{\pi}}(s) + \gamma \sum_{s'} P_{\bm{\pi}}(s'|s)  \bm{V}_{\bm{\pi}}(s') \; \text{ for all } s\in \mathcal{S}
	\end{aligned}  
	\end{equation}
	Where $\bm{r}_{\bm{\pi}}(s)$ is the expected immediate reward at state $s$ under ${\bm{\pi}}$ and $P_{\bm{\pi}}(s'|s)$ is the transition probability of moving from $s$ to $s'$ under ${\bm{\pi}}$. Notice that the transition probability is a combination of policy and the environment dynamics. We can write $\bm{V}_{\bm{\pi}},\bm{r}_{\bm{\pi}}, P_{\bm{\pi}}$ in matrix form
	\begin{equation} \label{original matrix definitions}
	\bm{V}_{\bm{\pi}} \!=\!
	\begin{bmatrix}
	\bm{V}_{\bm{\pi}}(s_1) \\
	\bm{V}_{\bm{\pi}}(s_2)  \\
	\vdots\\
	\bm{V}_{\bm{\pi}}(s_n) 
	\end{bmatrix},
	\;
	\bm{r}_{\bm{\pi}} \!=\!
	\begin{bmatrix}
	\bm{r}_{\bm{\pi}}(s_1) \\
	\bm{r}_{\bm{\pi}}(s_2) \\ 
	\vdots \\
	\bm{r}_{\bm{\pi}}(s_n) 
	\end{bmatrix}
	\!=\!
	\begin{bmatrix}
	\sum\limits_a {\bm{\pi}}(a|s_1) r(s_1,a) \\
	\sum\limits_a {\bm{\pi}}(a|s_2) r(s_2,a)  \vspace{-2mm}\\
	\vdots\\
	\sum\limits_a {\bm{\pi}}(a|s_n) r(s_n,a)
	\end{bmatrix}, 
	\;
	P_{\bm{\pi}} \!=\!
	\begin{bmatrix}
	P_{\bm{\pi}}(s_1|s_1) & P_{\bm{\pi}}(s_2|s_1)& \cdots & P_{\bm{\pi}}(s_n|s_1)\\
	P_{\bm{\pi}}(s_1|s_2) & P_{\bm{\pi}}(s_2|s_2)& \cdots & P_{\bm{\pi}}(s_n|s_2)  \\
	\vdots & \vdots & \ddots & \vdots\\
	P_{\bm{\pi}}(s_1|s_n) & P_{\bm{\pi}}(s_2|s_n)& \cdots & P_{\bm{\pi}}(s_n|s_n) 
	\end{bmatrix}
	\end{equation}
	Then the Bellman equation can be rewritten in matrix form
	$$\bm{V}_{\bm{\pi}} = \bm{r}_{\bm{\pi}} + \gamma P_{\bm{\pi}} \bm{V}_{\bm{\pi}}$$
	Assume that $P_{\bm{\pi}}$ and $\bm{r}_{\bm{\pi}}$ are given and $I-\gamma P_{\bm{\pi}}$ is nonsingular, then
	\begin{equation}\label{Original Bellman Equation}
	\bm{V}_{\bm{\pi}} = (I-\gamma P_{\bm{\pi}})^{-1} \bm{r}_{\bm{\pi}}
	\end{equation}
	For later uses, we define the vector $\bm{Q}_\pi=[\bm{Q}(s_1,\cdot)^T, \bm{Q}(s_2,\cdot)^T, \ldots, \bm{Q}(s_n,\cdot)^T]^T$.
	\subsection{Problem}
	
	One problem with \eqref{Original Bellman Equation} is that $P_{\bm{\pi}}$ and $\bm{r}_{\bm{\pi}}$ are both dependent on ${\bm{\pi}}$, which means every time when ${\bm{\pi}}$ changes, we need to reconstruct them. In addition, although $\bm{V}_{\bm{\pi}}$ is a function of ${\bm{\pi}}$, we cannot write the function in terms of $\bm{\pi}$ explicitly. 
	What we desire is to have
	\begin{equation} \label{Ideal Objective}
	\bm{V}_{\bm{\pi}} = f({\bm{\pi}} | P,r) \quad \text{or} \quad \bm{V}_{{\bm{\pi}}_{\theta}} = f({\bm{\pi}}_{\theta} | P,r)
	\end{equation}
	%\textcolor{red}{(What does $f(a|b)$ means? If it is only a function of $\pi$, why not write it as $f(\pi)$? (Because it is conditioned on the environment $P,r$.)1. For a general function, this notation is not a standard one. An explanation is needed, unless people in machine learning has used it.  2. This would mean $\bm\pi$ depends on $P$ and $r$. Is it true?) (1. Yes, it's a common way. 2. No, this means given the value of $P,r$, the output of the function depends on the value of variable $\bm{\pi}$)}
	
	where $\bm{\pi}_\theta$ is a parameterized policy, instead of 
	$\bm{V}_{\bm{\pi}} = f({\bm{\pi}} | P_{\bm{\pi}},\bm{r}_{\bm{\pi}})$
	or $\bm{V}_{{\bm{\pi}}_{\theta}} = f({\bm{\pi}}_{\theta} | P_{{\bm{\pi}}_{\theta}},r_{{\bm{\pi}}_{\theta}})$.
	Then, we can directly write our objective as a function of ${\bm{\pi}}$, which can make optimization more straightforward and efficient, just as follows,
	\begin{equation}\label{Ideal Optimization}
	\underset{{\bm{\pi}}}{\text{argmax}} \; \bm{V}_{\bm{\pi}} = f({\bm{\pi}} | P,r) \quad \text{or} \quad  \underset{\theta}{\text{argmax}} \; \bm{V}_{{\bm{\pi}}_{\theta}} = f({\bm{\pi}}_{\theta} | P,r)
	\end{equation}
	and solve it via gradient descent or other techniques. The notations from \cite{wang2007dual} is helpful to construct \eqref{Ideal Objective}\eqref{Ideal Optimization}. We introduce the formulation in the following sections and try to rewrite TRPO \cite{schulman2015trust} with it.
	
	\section{Notation and Preliminary}
	
	\subsection{Notation and Properties}
	\label{notation}
	For simplicity, we assume both the number of states in $\cal S$ and the number of actions in $\cal A$
	are finite. We define the notations as follows \cite{wang2007dual}:
	\begin{itemize}
		\item $|\cal S|$ and $|\cal A|$ denote the number of states in $\cal S$ 
		and the number of actions in $\cal A$, respectively. 
		\item  $P \in \mathbb{R}^{|S||A| \times |S|}$ is a transition matrix whose entries are $ P_{\left(s a, s^{\prime}\right)} = p\left(s^{\prime} | s, a\right)$, 
		where $ p\left(s^{\prime} | s, a\right) \geq 0, \; \sum_{s^{\prime}} p\left(s^{\prime} | s, a\right)=1,$ for all $s$ and $a$, %specify the conditional probability of transition into state $s'$ starting from state $s$ and taking action $a$, 
		\ie, 
		\begin{equation} \label{eq:Pproperty}
		P\bm{1}_{|S|}=\bm{1}_{|S||A|}, \quad \bm{1}_{|S|}=[1,1,\ldots,1]^T\in \mathbb{R}^{|S|\times 1} 
		\end{equation}
		%For example, for ${\cal S}=\{s_1,s_2,s_3\}$ and ${\cal A}=\{a_1,a_2\}$,
		%$$
		%P=\begin{bmatrix} 
		%p(s_1|s_1,a_1) & p(s_2|s_1,a_1) & p(s_3|s_1,a_1) \\
		%p(s_1|s_1,a_2) & p(s_2|s_1,a_2) & p(s_3|s_1,a_2) \\
		%p(s_1|s_2,a_1) & p(s_2|s_2,a_1) & p(s_3|s_2,a_1) \\
		%p(s_1|s_2,a_2) & p(s_2|s_2,a_2) & p(s_3|s_2,a_2) \\
		%p(s_1|s_3,a_1) & p(s_2|s_3,a_1) & p(s_3|s_3,a_1) \\
		%p(s_1|s_3,a_2) & p(s_2|s_3,a_2) & p(s_3|s_3,a_2)
		%\end{bmatrix}
		%$$
		%where the sum of the elements in each row is 1.
		
		\item  ${\bm{\pi}} \in \mathbb{R}^{|S||A| \times 1}$ is a stationary policy, whose entries are ${\bm{\pi}}_{(s a)} = \bm{\pi}{(a|s)}$, where $\sum_{a} {\bm{\pi}}_{(s a)}=1$. It is convenient to rewrite the policy as a matrix $\Pi \in \mathbb{R}^{|S|\times |S||A|}$, where $\Pi_{\left(s, s^{\prime} a\right)}={\bm{\pi}}_{(s a)} \text { if } s^{\prime}=s$, otherwise 0, \ie{}
		\begin{equation} \label{policy}
		\Pi = \text{diag}(\bm\pi(\cdot | s_1)^T, \cdots, \bm\pi(\cdot | s_{|S|})^T), \; 
		\bm\pi(\cdot | s_i)^T \in \mathbb{R}^{1 \times |A|}.
		\end{equation}
		$\Xi \in \mathbb{R}^{|S|\times|S||A|}$ is an auxiliary (marginalization) matrix defined by
		\begin{equation} \label{auxiliary matrix}
		\Xi =  \text{diag}(\bm{1}_{|A|}^T , \cdots, \bm{1}_{|A|}^T)
		\end{equation}
		%$$ \Xi = 
		%\left( \begin{array} { cccc } 
		%\bm{1}_{|A|}^T & & & \\ 
		%& \bm{1}_{|A|}^T & & \\ 
		%& & \ddots& \\ 
		%& & & \bm{1}_{|A|}^T \\ 
		%\end{array} \right)$$
		%where $\bm{1}_{|A|}^T \in \mathbb{R}^{1\times|A|} $ consists of all 1s. 
		It is easy to verify that
		\begin{equation} \label{auxiliary matrix property}
		\Pi \bm{1}_{|S||A|} = |A| \bm{1}_{|S|}, \quad
		\Pi \Xi^T = I_{|S|}
		\end{equation}
		%That is, $\Pi$ is a sparse matrix built by placing $|S|$ row blocks of length $|A|$  in a block diagonal fashion, where each row block gives the conditional distribution over actions specified by ${\bm{\pi}}$ in a particular state $s$. Although this representation of $ \Pi$ might appear unnatural, we find it in fact extremely convenient in our research: from this definition, one can quickly verify that the $|S| \times|S|$ matrix product  $\Pi P $ gives the state to state transition probabilities induced  by the policy ${\bm{\pi}}$  in the environment P, and the $|S||A| \times|S||A|$ matrix product $P \Pi$ gives the state-action to state-action transition probabilities induced by policy ${\bm{\pi}}$ in the environment $P$.
		%$\bm{\eta}^{\top}=(1-\gamma) \boldsymbol{\mu}^{\top}+(1-\gamma) \boldsymbol{\mu}^{\top} \sum\limits_{i=1}^{\infty} \gamma^{i}(\Pi P)^{i}=(1-\gamma) \boldsymbol{\mu}^{\top}+\gamma \bm{\eta}^{\top} \Pi P$
		We can easily reconstruct the state-to-state transition matrix by $P_{\bm{\pi}} = \Pi P \in \mathbb{R}^{|S| \times|S|}$, and state-action-to-state-action transition matrix by $P_{\bm{\pi}}'= P \Pi \in \mathbb{R}^{|S||A| \times|S||A|}$.
		From \eqref{eq:Pproperty} and \eqref{auxiliary matrix property},
		$$
		P_{\bm{\pi}}\bm{1}_{|S|}=\Pi P \bm{1}_{|S|}=\Pi \bm{1}_{|S||A|} =  \bm{1}_{|S|}
		$$
		$$
		P_{\bm{\pi}}'\bm{1}_{|S||A|}= P \Pi \bm{1}_{|S||A|} = P \bm{1}_{|S|} = \bm{1}_{|S||A|}
		$$

		\item $\bm{\rho}_0  \in \mathbb{R}^{|S| \times 1} $ is the initial state distribution. $\bm{\mu_{\pi}} \in \mathbb{R}^{|S| \times 1} $ is the steady-state distribution of $\pi$ given environment $P$ which satisfies $\bm{\mu_{\pi}}^T \Pi P = \bm{\mu_{\pi}}^T$. $\bm{\rho}_{\bm{\pi}}^T = \bm{\rho}_0^T \sum_{i=0}^\infty \gamma^i (\Pi P)^i = \bm{\rho}_0^T (I - \gamma \Pi P)^{-1}$ is the discounted visitation frequency vector under $\bm{\pi}$ and 
		$$
		\textstyle{
			\bm{\rho}_{\bm{\pi}}^T \bm{1} 
			= \bm{\rho}_0^T  \sum_{i=0}^\infty \gamma^i (\Pi P)^i  \bm{1}
			=\bm{\rho}_0^T (\sum_{i=0}^\infty  \gamma^i) \bm{1}
			=1/(1-\gamma)
		}
		$$
		
		\item $\bm{r} \in \mathbb{R}^{|S||A|\times 1} $ is the average reward vector whose entries are $\bm{r}_{(sa)} = r(s,a) = E[r|s,a] = \sum_{s',r}  r \cdot p(s',r|s,a)$, which specify the average reward obtained when taking action $a$ in state $s$.
		\begin{comment}
		For the case: ${\cal S}=\{s_1,s_2,s_3\}$ and ${\cal A}=\{a_1,a_2\}$,
		$$
		\bm{r}= [r(s_1,a_1),r(s_1,a_2),r(s_2,a_1),r(s_2,a_2),r(s_3,a_1),r(s_3,a_2)]^T
		$$
		\end{comment}
		From the definition of $\bm{r}_{\bm\pi}$ in \eqref{original matrix definitions}
		and $\Pi$ in \eqref{policy}, it is easy to see that
		\begin{equation} \label{eq:rpipir}
		\bm{r}_{\bm\pi} = \Pi \bm{r}
		\end{equation}
		
		%\item 
		%$M=(1-\gamma) \sum\limits_{i=0}^{\infty} \gamma^{i}(\Pi P)^{i}$
		%$M=(1-\gamma) I+(1-\gamma) \sum\limits_{i=1}^{\infty} \gamma^{i}(\Pi %P)^{i}=(1-\gamma) I+\gamma \Pi P M$
		%$M = \left( \begin{array} { c } { m _ { 1 } ^ { \top } } \\ { m _ { 2 } ^ { \top } } \\ { m _ { 3 } ^ { \top } } \\ { \vdots } \\ { m _ { | S | } ^ { \top } } \end{array} \right) = \left( \begin{array} { c } { ( 1 - \gamma ) \bm { e } _ { 1 } ^ { \top } + \gamma \Pi P m _ { 1 } ^ { \top } } \\ { ( 1 - \gamma ) \bm { e } _ { 2 } ^ { \top } + \gamma \Pi P m _ { 2 } ^ { \top } } \\ { ( 1 - \gamma ) \bm { e } _ { 3 } ^ { \top } + \gamma \Pi P m _ { 3 } ^ { \top } } \\ { \vdots } \\ { ( 1 - \gamma ) \bm { e } _ { | S | } ^ { \top } + \gamma \Pi P m _ { | S | } ^ { \top } } \end{array} \right)$
		%$M \bm { 1 } = \bm { 1 }$
		%$\mathcal { O } M = ( 1 - \gamma ) I + \gamma M \Pi P$
		%$\mathcal{O}_{\bm{Q}} = \bm { r } + \gamma P \Pi \bm { Q }$
		%\item 
		%$\bm{d}^{ \top } = (1 - \gamma ) \boldsymbol{\mu}^{\top} \sum\limits_{i = 0}^{\infty} \gamma^{i} (P \Pi)^{i}$
		%where $\nu $ is the initial distribution over the state-action pairs, "whose dimension is $| S | | A | \times 1$
		%$\bm { d } ^ { \top } = ( 1 - \gamma ) \bm{\mu}^{ \top } + \gamma \bm { d } ^ { \top } P \Pi$
		%\item 
		%Consider the following definition for an $|S||A| \times|S||A|$ matrix
		%$H=(1-\gamma) \sum\limits_{i=0}^{\infty} \gamma^{i}(P \Pi)^{i} = (1-\gamma) I+\gamma P \Pi H$
		%\item 
		%For a given policy $\Pi$,  the on-policy operator $\mathcal{O}$  is defined as  \\ $\mathcal{O H}=(1-\gamma) I+\gamma P \Pi H$
	\end{itemize}
	
	Now we can rewrite the Bellman equation. 
	From \eqref{original matrix definitions}, we have
	\begin{equation} \label{value function}
	\textstyle{
		\bm{V}_{\bm{\pi}} = (I-\gamma P_{\bm{\pi}})^{-1} \bm{r}_{\bm{\pi}} = (I-\gamma \Pi P)^{-1} \Pi \bm{r} = \sum_{i=0}^{\infty} \gamma^{i}(\Pi P)^{i} \Pi \bm{r} = \Pi \bm{r}+\gamma \Pi P \bm{V}_{\bm{\pi}}
	}
	\end{equation}
	Similarly, we have $\bm{Q}_{\bm{\pi}} = \sum\limits_{i = 0}^{\infty} \gamma^{ i } ( P \Pi )^{i} \bm{r} = \bm{r} + \gamma P \Pi \bm {Q}_{\bm{\pi}}$.
	%$\bm{V}=\Pi \bm{r}+\sum\limits_{i=1}^{\infty} \gamma^{i}(\Pi P)^{i} \Pi \bm{r}=\Pi \bm{r}+\gamma(\Pi P) \sum\limits_{i=0}^{\infty} \gamma^{i}(\Pi P)^{i} \Pi \bm{r}$. %For a given policy   $\Pi$, the on-policy operator $\mathcal{O}$  is defined as $\mathcal{O}_{\bm{V}}=\Pi(\bm{r}+\gamma P \bm{V}) $
	%The main problem is to compute an optimal policy given either a complete specification of the environmental model $P$ and $r$ (the “planning problem”), or limited access to the environment through observed states and rewards and the ability to select actions to cause further state transitions (the “learning problem”). The planning problem is normally tackled by linear programming or dynamic programming methods, whereas the learning problem is solved by reinforcement learning methods.
	One can easily check that the relations between $\bm{V}_{\bm{\pi}}$ and $\bm{Q}_{\bm{\pi}}$ are as follows,
	\begin{equation} \label{value function relation}
	\textstyle{
		\bm{V}_{\bm{\pi}} = \Pi \bm{Q}_{\bm{\pi}}, \; \bm{Q}_{\bm{\pi}} = \sum_{i = 0}^{\infty} \gamma^{i} (P \Pi )^{i} \bm{r} = \bm {r} + \gamma P [\sum_{i = 0}^{\infty} \gamma^{i} (P \Pi )^{i} \Pi \bm{r}]  = \bm {r} + \gamma P \bm{V}_{\bm{\pi}}
	}
	\end{equation}
	
	%Some Lemmas and corollaries
	%\begin{itemize}
	%\item Lemma 2 $\mathrm{c}^{\top} \bm{1}=1$
	%$(1-\gamma) \boldsymbol{\mu}^{\top} \bm{V}=\bm{c}^{\top} \Pi \bm{r}$
	%\item 
	%Lemma 5  $\bm { c } ^ { \top } = \boldsymbol { \mu } ^ { \top } M$
	%\item Theorem  6 $( 1 - \gamma ) \bm { V } = M \Pi \bm { r }$
	%\item Lemma 7 $\mathrm { d } ^ { \top } \bm { 1 } = 1$
	%\item Lemma 8 $ ( 1 - \gamma ) \nu ^ { \top } \bm { Q } = \bm { d } ^ { \top } \bm { r }$
	%\item Lemma  9 $\mathrm{H} 1=1$
	%\item Lemma 10 $\mathrm{d}^{\top}=\nu^{\top} H$
	%\item Relationship between Primal and Dual Representations.  \\
	%Theorem 11 $(1-\gamma) \bm{Q}=H \bm{r}$
	%\item Lemma 12 $\mathrm{V}=\Pi \mathrm{Q}$
	%\item Lemma  13 $\mathrm{M\Pi}=\mathrm{\Pi} H$
	%\item 
	%$\bm {Q} = \sum\limits_{i = 0}^{\infty} \gamma^{i} (P \Pi )^{i} \bm{r} = \bm {r} + \sum\limits_{i = 1}^{\infty} \gamma^{i} (P \Pi )^{i} \bm{r} = \bm {r} + \gamma P [\sum\limits_{i = 0}^{\infty} \gamma^{i} (P \Pi )^{i} \Pi \bm{r}]  = \bm {r} + \gamma P V $
	%\end{itemize}
	\subsection{A Recap of TRPO}
	We do a simple recap of TRPO \cite{schulman2015trust} in this subsection. We will show how these results can easily be obtained when written in matrix form and how to extend them.
	
	For policy-based reinforcement learning algorithms, if we use policy gradient with a fixed learning rate to update the policy, it always happens that the learning rate is sometimes too large that we will get a worse policy. This oscillation makes the training unstable and the convergence slow. So we want to find a way that our policy is guaranteed to improve after each policy update even the improvement might be quite small at some time. The objective $\eta({\bm{\pi}})$ is defined as follows, %Suppose we have a policy ${\bm{\pi}}$ and a new policy $\tilde{{\bm{\pi}}} = {\bm{\pi}} + d {\bm{\pi}}$, %(or through parameter update $\theta = \theta_{old} + d \theta$), 
	\begin{equation} \label{eta}
	\eta({\bm{\pi}}) = \mathbb{E}_{s_{0}, a_{0}, \ldots} \left[{\textstyle \sum_{t=0}^{\infty}} \gamma^{t} r\left(s_{t}\right)\right], \text{ where }  {s_{0} \sim \bm{\rho}_{0}\left(s_{0}\right), a_{t} \sim {\bm{\pi}}\left(a_{t} | s_{t}\right), s_{t+1} \sim p\left(s_{t+1} | s_{t}, a_{t}\right)}
	\end{equation}
	%we want to write $\eta(\tilde{{\bm{\pi}}})$ as 
	%$$\eta(\tilde{{\bm{\pi}}}) = \eta({\bm{\pi}}) + \epsilon({\bm{\pi}}, \bm{\pi} + d {\bm{\pi}}).$$
	%Then, if $\epsilon({\bm{\pi}},\bm{\pi} + d {\bm{\pi}}) \geq 0$, we will have a guarantee that $\eta(\tilde{{\bm{\pi}}}) \geq \eta({\bm{\pi}})$.%(or $\eta({\bm{\pi_\theta}}) \geq \eta({\bm{\pi_{\theta_{old}}}})$ ).
	With the definition of the advantage function $\bm{A}_{\bm{\pi}}(s,a)$:
	\begin{equation} \label{advantage}
	\bm{A}_{\bm{\pi}}(s, a) = \bm{Q}_{\bm{\pi}}(s, a) - \bm{V}_{\bm{\pi}}(s)
	\end{equation}
	%& Q_{{\bm{\pi}}}\left(s_{t}, a_{t}\right)=\mathbb{E}_{s_{t+1}, a_{t+1}, \ldots}\left[\sum_{l=0}^{\infty} \gamma^{l} r\left(s_{t+l}\right)\right], V_{{\bm{\pi}}}\left(s_{t}\right)=\mathbb{E}_{a_{t}, s_{t+1}, \ldots}\left[\sum_{l=0}^{\infty} \gamma^{l} r\left(s_{t+l}\right)\right], 
	%where $a_{t} \sim {\bm{\pi}}\left(a_{t} | s_{t}\right), s_{t+1} \sim P\left(s_{t+1} | s_{t}, a_{t}\right) \text { for } t \geq 0$, we have
	$\eta(\tilde{{\bm{\pi}}})$ can be split into two parts,
	\begin{equation}\label{eta relation}
	\eta(\tilde{\bm{\pi}}) = \eta(\bm{\pi}) + \mathbb{E}_{s_{0}, a_{0}, \cdots \sim \tilde{\bm{\pi}}} \left[\textstyle{\sum_{t=0}^{\infty}} \gamma^{t} \bm{A}_{\bm{\pi}} \left(s_{t}, a_{t}\right)\right] 
	= \eta(\bm{\pi}) + \textstyle{\sum_{s} \bm{\rho}_{\tilde{\bm{\pi}}}(s) \sum_{a}} \tilde{\bm{\pi}}(a|s) \bm{A}_{\bm{\pi}}(s, a)
	\end{equation}
	where $\bm{\rho}_{\tilde{\bm{\pi}}}(s) = \sum_{i=0}^\infty \gamma^i p\left( s_i=s\right | \tilde{\bm{\pi}}, P, \bm{\rho}_0^T)$ is the discounted visitation frequency of $s$ under $\tilde{\bm{\pi}}$.
	%$$\bm{\rho}_{{\bm{\pi}}}(s)=P\left(s_{0}=s\right)+\gamma P\left(s_{1}=s\right)+\gamma^{2} P\left(s_{2}=s\right)+\ldots$$
	%$$\begin{aligned} \eta(\tilde{{\bm{\pi}}}) &=\eta({\bm{\pi}})+\sum_{t=0}^{\infty} \sum_{s} P\left(s_{t}=s | \tilde{{\bm{\pi}}}\right) \sum_{a} \tilde{{\bm{\pi}}}(a | s) \gamma^{t} A_{{\bm{\pi}}}(s, a) = \eta({\bm{\pi}})+\sum_{s} \sum_{t=0}^{\infty} \gamma^{t} P\left(s_{t}=s | \tilde{{\bm{\pi}}}\right) \sum_{a} \tilde{{\bm{\pi}}}(a | s) A_{{\bm{\pi}}}(s, a) \\ & = \eta({\bm{\pi}})+\sum_{s} \bm{\rho}_{\tilde{{\bm{\pi}}}}(s) \sum_{a} \tilde{{\bm{\pi}}}(a | s) A_{{\bm{\pi}}}(s, a) \end{aligned}$$
	A local approximation to $\eta(\tilde{\bm{\pi}})$ is constructed as follows:
	\begin{equation} \label{local function approximation}
	L_{{\bm{\pi}}}(\tilde{{\bm{\pi}}})
	= \eta({\bm{\pi}})+ \textstyle{\sum_{s} \bm{\rho}_{{\bm{\pi}}}(s) \sum_{a}}
	\tilde{{\bm{\pi}}}(a | s) \bm{A}_{{\bm{\pi}}}(s, a)
	\end{equation}
	%$$\eta(\tilde{{\bm{\pi}}})=\eta({\bm{\pi}})+\mathbb{E}_{\tau \sim \tilde{{\bm{\pi}}}}\left[\sum_{t=0}^{\infty} \gamma^{t} A_{{\bm{\pi}}}\left(s_{t}, a_{t}\right)\right]$$
	It satisfies two important properties when $\bm{\pi}_\theta$ is parameterized by $\theta$:
	\begin{equation} \label{local approximation properties}
	L_{\bm{\pi}_{\theta_{0}}} \left({\bm{\pi}}_{\theta_{0}}\right) = \eta\left({\bm{\pi}}_{\theta_{0}}\right), \ \
	\left. \nabla_{\theta}  L_{\bm{\pi}_{\theta_{0}}} \left({\bm{\pi}}_{\theta}\right) \right|_{\theta=\theta_{0}} = \left. \nabla_{\theta} \eta\left( {\bm{\pi}}_{\theta} \right) \right|_{\theta=\theta_{0}} 
	\end{equation}
	%$$\begin{array}{c}{\eta\left({\bm{\pi}}_{\mathrm{new}}\right) \geq L_{{\bm{\pi}}_{\mathrm{old}}}\left({\bm{\pi}}_{\mathrm{new}}\right)-\frac{2 \epsilon \gamma}{(1-\gamma)^{2}} \alpha^{2}} \\ {\text { where } \epsilon=\max _{s}\left|\mathbb{E}_{a \sim {\bm{\pi}}^{\prime}(a | s)}\left[A_{{\bm{\pi}}}(s, a)\right]\right|}\end{array}$$
	Given the total variation divergence for two discrete distributions $D_{T V}(p \| q) = \frac{1}{2}  \sum_{i}\left|p_{i}-q_{i}\right|$, we define $D_{\mathrm{TV}}^{\max }(\bm{\pi}, \tilde{\bm{\pi}}) = \max_{s} D_{\mathrm{TV}} \left(\bm{\pi}( \cdot | s) \| \tilde{\bm{\pi}}(\cdot | s) \right)$. 
	It can be proved that
	\begin{equation}
	\eta \left(\tilde{\bm{\pi}}\right) \geq L_{\bm{\pi}} \left( \tilde{\bm{\pi}} \right) - \frac{4 \epsilon \gamma}{(1-\gamma)^{2}} \alpha^{2}, \text{ where } \alpha = D_{\mathrm{TV}}^{\max } \left({\bm{\pi}}, \tilde{\bm{\pi}} \right),\;  \epsilon = \max_{s, a} \left|\bm{A}_{\bm{\pi}}(s, a)\right|
	\end{equation}
	With $D_{\mathrm{TV}}(p \| q)^{2} \leq D_{\mathrm{KL}}(p \| q),$ we define $D_{\mathrm{KL}}^{\max }(\bm{\pi}, \tilde{\bm{\pi}})=\max _{s} D_{\mathrm{KL}}(\bm{\pi}(\cdot | s) \| \tilde{\bm{\pi}}(\cdot | s))$. 
	Then we have
	\begin{equation}  \label{inequality KL}
	\eta(\tilde{\bm{\pi}}) \geq  L_{\bm{\pi}}(\tilde{\bm{\pi}})-C D_{\mathrm{KL}}^{\max }(\bm{\pi}, \tilde{\bm{\pi}}), \text{ where } C=\frac{4 \epsilon \gamma}{(1-\gamma)^{2}} 
	\end{equation}
	Let $M_{i}(\bm{\pi}) = L_{\bm{\pi}_i}(\bm{\pi}) - C D_{\mathrm{KL}}^{\max }\left(\bm{\pi}_{i}, {\bm{\pi}} \right)$, we have
	$$ \eta\left({\bm{\pi}}_{i+1}\right) \geq M_{i}\left({\bm{\pi}}_{i+1}\right), \eta\left({\bm{\pi}}_{i}\right)=M_{i}\left({\bm{\pi}}_{i}\right) \Rightarrow \eta\left({\bm{\pi}}_{i+1}\right)-\eta\left({\bm{\pi}}_{i}\right) \geq M_{i}\left({\bm{\pi}}_{i+1}\right)-M_i\left({\bm{\pi}}_{i}\right) $$
	So if we define ${\bm{\pi}}_{i+1}$ as 
	${\bm{\pi}}_{i+1} = \underset{{\bm{\pi}}}{\text{argmax}}\; M_{i}\left({\bm{\pi}} \right) $, then
	\begin{equation} 
	\eta\left({\bm{\pi}}_{0}\right) \leq \eta\left({\bm{\pi}}_{1}\right) \leq \eta\left({\bm{\pi}}_{2}\right) \leq \cdots
	\end{equation}
	Therefore, $M_{i}(\bm{\pi})$ becomes a surrogate function that we want to maximize.
	%$$\eta(\tilde{{\bm{\pi}}})=\eta({\bm{\pi}})+\mathbb{E}_{\tau \sim \tilde{{\bm{\pi}}}}\left[\sum_{t=0}^{\infty} \gamma^{t} A_{{\bm{\pi}}}\left(s_{t}, a_{t}\right)\right]$$

	\subsection{Relations Between Several Distance Measures}
	The total variation distance between two probability measures $\mu$ and $\nu$ on a sigma-algebra $\mathcal{F}$ of subsets of the sample space $\Omega$ is defined as $\delta(\mu, \nu)=\sup_{ A\in \mathcal{F}}\left|\mu(A)-\nu(A)\right|.$
	%Informally, this is the largest possible difference between the probabilities that the two probability distributions can assign to the same event.
	The total variation distance is related to the Kullback–Leibler divergence by Pinsker's inequality:
	$$
	\delta(\mu, \nu) \le \sqrt{\frac{1}{2} D_{\mathrm{KL}}(\mu , \nu)}
	$$
	And from \cite{levin2017markov}, we have
	$$
	\delta(\mu, \nu)=\frac12\|\mu -\nu\|_1=\frac12\sum_{\omega\in\Omega}|\mu(\omega)-\nu(\omega)|
	$$
	Thus,
	$$ 
	\|\mu -\nu\|_1^2 \leq 2 D_{\mathrm{KL}}(\mu , \nu)
	$$

	\section{TRPO in Matrix Form}
	In this section, we first write TRPO in matrix form and we will see there are more ways to find the local approximation as \eqref{local function approximation}. We derive several interesting properties of these approximations. The norm $\norm{\cdot}$ we use in this section is $1$-norm. Note that $\Pi$ (matrix) is just a rewriting of $\bm{\pi}$ (vector). They represent the same policy. This is the same for parametrized $\Pi_\theta$ and $\bm{\pi}_\theta$.
	
	With notations introduced in Section \ref{notation}, $\eta$ can be written as 
	$$
	{\eta}(\bm{\pi}) = \bm{\rho}_0^{\top} \bm{V}_{\bm{\pi}} = \bm{\rho}_0^{\top} \sum\limits_{i=0}^{\infty} \gamma^{i}(\Pi P)^{i} \Pi \bm{r}
	$$
	From \eqref{advantage},
	we see that the advantage function $\bm{A}_{\bm{\pi}}$, the vector form of $\bm{A_\pi}(s,a)$, can be written as
	$$\bm{A}_{\bm{\pi}} = \bm{Q}_{\bm{\pi}} - \Xi^T \bm{V}_{\bm{\pi}}$$
	From \eqref{auxiliary matrix property},\eqref{value function} and \eqref{value function relation}, 
	it is easy to verify that $\Pi \bm{A}_{\bm{\pi}} = 0$. 
	Then
	\begin{equation}
	(I-\gamma P \Pi )^{-1} \bm{A}_{\bm{\pi}} = \sum\limits_{i = 0}^{\infty} \gamma^{ i } ( P \Pi )^{i} \bm{A}_{\bm{\pi}} = \bm{A}_{\bm{\pi}} + \sum\limits_{i = 1}^{\infty} \gamma^{ i } ( P \Pi )^{i} \bm{A}_{\bm{\pi}} = \bm{A}_{\bm{\pi}} \\
	\end{equation}
	Since  
	$ {\eta}(\tilde{\bm{\pi}})= \bm{\rho}_0^T \bm{V}_{\tilde{\bm{\pi}}}$
	amd $\eta(\bm{\pi}) = \bm{\rho}_0^T \bm{V}_{\bm{\pi}}$,
	%$$A_{\bm{\pi}} = Q_\Pi - \Xi^{T} \bm{V}_{\bm{\pi}}$$
	\eqref{eta relation} can easily be shown as follows,
	\begin{align*}
	&\textstyle{\mathbb{E}_{s_{0}, a_{0}, \cdots \sim \bm{\tilde{\pi}}} \left[\sum_{t=0}^{\infty} \gamma^{t} \bm{A}_{{\bm{\pi}}} \left(s_{t}, a_{t}\right)\right]} \\
	& = \bm{\rho}_0^T \tilde{\Pi} (I-\gamma P \tilde{\Pi})^{-1} \bm{A}_{\bm{\pi}} = \bm{\rho}_0^T \tilde{\Pi} (I-\gamma P \tilde{\Pi})^{-1} [\bm{Q}_{\bm{\pi}} - \Xi^{T} \bm{V}_{\bm{\pi}}] \\
	& = \bm{\rho}_0^T \tilde{\Pi} (I-\gamma P \tilde{\Pi})^{-1} [\bm{r} + \gamma P \bm{V}_{\bm{\pi}} - \Xi^{T} \bm{V}_{\bm{\pi}}]  \\  
	& =\textstyle{
		\bm{\rho}_0^T \tilde{\Pi} \sum\limits_{i = 0}^{\infty} \gamma^{i} (P \tilde{\Pi} )^{i} \bm {r} + \bm{\rho}_0^T \tilde{\Pi} \sum_{i = 0}^{\infty} \gamma^{i} (P \tilde{\Pi} )^{i} \gamma P \bm{V}_{\bm{\pi}} - \bm{\rho}_0^T \tilde{\Pi} \sum_{i = 0}^{\infty} \gamma^{i} (P \Pi )^{i} \Xi^{T} \bm{V}_{\bm{\pi}}
	} \\ 
	& = \textstyle{
		\bm{\rho}_0^T  \sum_{i = 0}^{\infty} \gamma^{i} (\tilde{\Pi} P)^{i} \tilde{\Pi} \bm {r} + \bm{\rho}_0^T \sum_{i = 0}^{\infty} \gamma^{i+1} (\tilde{\Pi} P )^{i+1} \bm{V}_{\bm{\pi}} - \bm{\rho}_0^T \sum_{i = 0}^{\infty} \gamma^{i} (\Pi P)^{i} \bm{V}_{\bm{\pi}}
	}\\ 
	& = \bm{\rho}_0^T \bm{V}_{\tilde{\bm{\pi}}} - \bm{\rho}_0^T \bm{V}_{\bm{\pi}} 
	= \eta(\tilde{\bm{\pi}}) - \eta(\bm{\pi})
	\end{align*}
	Thus \eqref{Ideal Objective} can be written in the following form
	$$
	\eta(\tilde{\bm{\pi}}) = \eta(\bm{\pi}) + \bm{\rho}_0^T \tilde{\Pi} (I-\gamma P \tilde{\Pi})^{-1} A_{\bm{\pi}} = \eta(\bm{\pi}) + \bm{\rho}_0^T (I-\gamma \tilde{\Pi} P)^{-1} \tilde{\Pi} A_{\bm{\pi}} 
	\equiv \eta(\bm{\pi}) + f_{\bm{\pi}}(\tilde{\bm{\pi}})
	$$
	It is easy to see that $f_{\bm{\pi}}(\bm{\pi}) = 0$. 
	Suppose $\tilde{\Pi} = \Pi + d \Pi\; (d \Pi \rightarrow 0 \text{ and } d \Pi \bm{1}_{|S||A|\times 1} = 0)$ 
	and if we want $\eta(\tilde{\bm{\pi}}) \geq \eta(\bm{\pi})$, we should have $f_{\bm{\pi}}(\tilde{\bm{\pi}}) = f_{\bm{\pi}}( \bm{\pi} + d \bm{\pi}) \geq 0$.
	Note that
	\begin{equation}
	\begin{aligned}
	f_{\bm{\pi}}(\tilde{\bm{\pi}}) 
	& = f_{\bm{\pi}}(\tilde{\bm{\pi}}) - f_{\bm{\pi}}( \bm{\pi} ) 
	= \left. df_{\bm{\pi}}(\tilde{\bm{\pi}}) \right|_{\tilde{\bm{\pi}} = \bm{\pi}} \\
	& =  \left. \mathrm{trace}\left( \bm{\rho}_0^T (I-\gamma \tilde{\Pi} P)^{-1} \gamma (d {\Pi}) P (I-\gamma \tilde{\Pi} P)^{-1} \tilde{\Pi} \bm{A}_{\bm{\pi}} + \bm{\rho}_0^T (I-\gamma \tilde{\Pi} P)^{-1} (d {\Pi}) \bm{A}_{\bm{\pi}} \right) \right|_{\tilde{\bm{\pi}} = \bm{\pi}}\\
	&= \mathrm{trace}\left( \bm{\rho}_0^T (I-\gamma \Pi P)^{-1} (d {\Pi}) \bm{A}_{\bm{\pi}} \right)  
	= \mathrm{trace}\left(\bm{A}_{\bm{\pi}}\bm{\rho}_0^T (I-\gamma \Pi P)^{-1} d {\Pi}\right)
	\end{aligned}
	\end{equation}
	Therefore,
	\begin{equation}
	\left(\left. \nabla_{\tilde{\Pi}} f_{\bm{\pi}}(\tilde{\bm{\pi}}) \right|_{\tilde{\bm{\pi}} = \bm{\pi}} \right)^T = \bm{A}_{\bm{\pi}}\bm{\rho}_0^T (I-\gamma \Pi P)^{-1}
	%\left. \left(\frac{\partial f_{\bm{\pi}}(\tilde{\bm{\pi}}) }{\partial \tilde{{\Pi}}}\right)^T \right|_{\tilde{\bm{\pi}} = \bm{\pi}} = \bm{A}_{\bm{\pi}}\bm{\rho}_0^T (I-\gamma \Pi P)^{-1}
	\end{equation}
	\begin{comment}
	If we have a parameterized policy $\bm{\pi}_\theta$, \textcolor{red}{(Is $\theta$ regarded as a scalar or a vector? If it's the latter, use $\bm{\theta}$)}which can be rewritten in matrix form as $\Pi_\theta$. 
	\textcolor{red}{(1. Why do you use both $\bm{\pi}_\theta$ and $\Pi_\theta$ in the following?
	2. What's the relation between $\Pi_\theta$ and $\Pi$ defined before - I am afraid the notation is confusing)}
	Then,
	\begin{equation}
	\begin{aligned}
	\left. \nabla_{\theta} \eta\left({\bm{\pi}}_{\theta} \right) \right|_{\theta=\theta_{0}} &= \left. \nabla_{\theta} \eta(\bm{\pi}_{\theta_0}) + \nabla_{\theta} \bm{\rho}_0^T (I-\gamma \Pi_\theta P)^{-1} \Pi_\theta \bm{A}_{\bm{\pi_{\theta_0}}} \right|_{\theta=\theta_{0}} \\
	& = \left. \bm{\rho}_0^T (I-\gamma \Pi_\theta P)^{-1} \gamma (\nabla_{\theta} \Pi_\theta) P (I-\gamma \Pi_\theta P)^{-1} \Pi_\theta \bm{A}_{\bm{\pi_{\theta_0}}} + \bm{\rho}_0^T (I-\gamma \Pi_\theta P)^{-1} (\nabla_{\theta} \Pi_\theta) \bm{A}_{\bm{\pi_{\theta_0}}} \right|_{\theta=\theta_{0}}\\
	& = \bm{\rho}_0^T (I-\gamma \Pi_{\theta_0} P)^{-1} \left(\left. \nabla_{\theta} \Pi_\theta\right|_{\theta=\theta_{0}} \right) \bm{A}_{\bm{\pi_{\theta_0}}} 
	\end{aligned}
	\end{equation}
	\textcolor{red}{(Is $\nabla$ for gradient? Didn't you use $\partial/\partial$ before?
	If $\Pi_\theta$ is a matrix, what is $\nabla_{\theta} \Pi_\theta$?)}
	\end{comment}
	There are 6 ways to set the approximation function as \eqref{local function approximation}:
	\begin{align*}
	L^1_{\bm{\pi}}(\tilde{\bm{\pi}}) & = \eta(\bm{\pi}) + \bm{\rho}_0^T \tilde{\Pi} (I-\gamma P \tilde{\Pi})^{-1} \bm{A}_{\bm{\pi}} = \eta(\bm{\pi}) + \bm{\rho}_0^T (I-\gamma \tilde{\Pi} P)^{-1} \tilde{\Pi} \bm{A}_{\bm{\pi}} \ \ (\text{policy gradient}) \\
	L^2_{\bm{\pi}}(\tilde{\bm{\pi}}) &= \eta(\bm{\pi}) + \bm{\rho}_0^T \Pi (I-\gamma P \tilde{\Pi})^{-1} \bm{A}_{\bm{\pi}} \\
	L^3_{\bm{\pi}}(\tilde{\bm{\pi}}) &= \eta(\bm{\pi}) + \bm{\rho}_0^T \tilde{\Pi} (I-\gamma P \Pi)^{-1} \bm{A}_{\bm{\pi}} \\
	L^4_{\bm{\pi}}(\tilde{\bm{\pi}}) &= \eta(\bm{\pi}) + \bm{\rho}_0^T (I-\gamma \Pi P)^{-1} \tilde{\Pi} \bm{A}_{\bm{\pi}} \ \  (\text{TRPO})\\
	L^5_{\bm{\pi}}(\tilde{\bm{\pi}}) &= \eta(\bm{\pi}) + \bm{\rho}_0^T (I-\gamma \tilde{\Pi} P)^{-1} \Pi \bm{A}_{\bm{\pi}} = \eta(\bm{\pi}) \; (\text{trivial})\\
	L^6_{\bm{\pi}}(\tilde{\bm{\pi}}) &= \eta(\bm{\pi}) + \bm{\rho}_0^T \Pi (I-\gamma P \Pi)^{-1} \bm{A}_{\bm{\pi}} = \eta(\bm{\pi}) + \bm{\rho}_0^T (I-\gamma \Pi P)^{-1} \Pi \bm{A}_{\bm{\pi}} = \eta(\bm{\pi}) \ \  (\text{trivial})
	\end{align*}
	We will discuss $L^2_{\bm{\pi}}(\tilde{\bm{\pi}}), L^3_{\bm{\pi}}(\tilde{\bm{\pi}}) \text{ and } L^4_{\bm{\pi}}(\tilde{\bm{\pi}})$ in the following subsections.
	%\subsection{Approximation Function $L^1_{\bm{\pi}}(\tilde{\bm{\pi}})$}
	%We need to estimate 
	%\begin{align*}( I-\gamma P{\Pi} )^{-1}  (Q_{\bm{\pi}} - \Xi^T \bm{V}_{\bm{\pi}}) &= (I-\gamma P{\Pi})^{-1}  ((I- \gamma P \Pi)^{-1} \bm { r } -\Xi^T (I- \gamma \Pi P )^{-1} \Pi \bm { r }) ) \\&= (I-\gamma P{\Pi})^{-1}  ((I- \gamma P \Pi)^{-1} \bm { r } -\Xi^T \Pi(I- \gamma P \Pi  )^{-1}  \bm { r }))\\&= (I-\gamma P{\Pi})^{-2} r - \sum\limits_{i=0}^{\infty} \gamma^{i}(P \Pi)^{i} \Xi^T (\Pi  (I- \gamma P \Pi)^{-1} r)\\& = (I-\gamma P{\Pi})^{-2} r - (\Xi^T + \gamma P + \gamma^{2} P \Pi P +   \gamma^{3} P \Pi P \Pi P + \cdots) (\Pi  (I- \gamma P \Pi)^{-1} r)\\ & = (I-\gamma P{\Pi})^{-2} r - (\Xi^T \Pi -I + I + \gamma P \Pi  + \gamma^{2} P \Pi P \Pi  +   \gamma^{3} P \Pi P \Pi P \Pi + \cdots)  (I- \gamma P \Pi)^{-1} r\\ & = (I-\gamma P{\Pi})^{-2} r - (\Xi^T \Pi -I + (I-\gamma P{\Pi})^{-1})  (I- \gamma P \Pi)^{-1} r\\& = (I - \Xi^T \Pi)  (I- \gamma P \Pi)^{-1} r\\ & = (I - \Xi^T \Pi)  Q_{\bm{\pi}} = Q_{\bm{\pi}} - \Xi^T \bm{V}_{\bm{\pi}} = A_{\bm{\pi}}
	%&= (I-\gamma P{\Pi})^{-1} (I - \Xi^T \Pi)  (I- \gamma P \Pi)^{-1} \bm {r}\\
	%\end{align*}
	
	\subsection{Approximation Function $L^2_{\bm{\pi}}(\tilde{\bm{\pi}})$}
	For $L^2_{\bm{\pi}}(\tilde{\bm{\pi}}) = \eta(\bm{\pi}) + \bm{\rho}_0^T {\Pi} (I-\gamma P \tilde{{\Pi}})^{-1} \bm{A}_{\bm{\pi}}$, it is easy to see that $L^2_{\bm{\pi}}(\bm{\pi}) = 0$.
	%If we set $$L_{\bm{\pi}}(\tilde{\bm{\pi}}) = \eta({\Pi}) + \bm{\rho}_0^T {\Pi} (I-\gamma P \tilde{{\Pi}})^{-1} \bm{A}_{\bm{\pi}} = \eta(\bm{\pi}) + f_{\bm{\pi}}(\tilde{\bm{\pi}})$$
	For parametrized $\bm{\pi}_\theta$, to test \eqref{local function approximation}, we have 
	\begin{equation}
	\begin{aligned}
	\left. d L^2_{\bm{\pi_{\theta_0}}}\left({\bm{\pi}}_{\theta} \right) \right|_{\theta=\theta_{0}} &= \left. d \bm{\rho}_0^T \Pi_{\theta_0} (I-\gamma P \Pi_\theta)^{-1}  \bm{A}_{\bm{\pi_{\theta_0}}} \right|_{\theta=\theta_{0}} \\
	& = \mathrm{trace} \left( \left. \bm{\rho}_0^T \Pi_{\theta_0} (I-\gamma P \Pi_\theta)^{-1} \gamma P (d \Pi_\theta) (I-\gamma P \Pi_\theta)^{-1} \bm{A}_{\bm{\pi_{\theta_0}}} \right) \right|_{\theta=\theta_{0}}\\
	& = \mathrm{trace} \left( \left. \bm{A}_{\bm{\pi_{\theta_0}}} \bm{\rho}_0^T \gamma \Pi_{\theta_0} P (I-\gamma \Pi_{\theta_0} P)^{-1}  \left( d \Pi_\theta \right)  \right) \right|_{\theta=\theta_{0}}
	\end{aligned}
	\end{equation}
	It is easy to see
	
	\begin{equation}
	\left( \left. \nabla_{\theta} L^2_{\bm{\pi_{\theta_0}}}\left({\bm{\pi}}_{\theta} \right) \right|_{\theta=\theta_{0}} \right)^T = \left. \bm{A}_{\bm{\pi_{\theta_0}}} \bm{\rho}_0^T \gamma \Pi_{\theta_0} P (I-\gamma \Pi_{\theta_0} P)^{-1} \frac{\partial \Pi_{\theta}}{\partial \theta}\right|_{\theta=\theta_{0}} \neq  \left( \left. \nabla_{\theta} \eta\left({\bm{\pi}}_{\theta} \right) \right|_{\theta=\theta_{0}} \right)^T
	\end{equation}
	which means $L^2_{\bm{\pi}}(\tilde{\bm{\pi}})$ does not match $\eta(\bm{\pi})$ to the first order. To calculate the difference between $L^2_{\bm{\pi}}(\tilde{\bm{\pi}})$ and $\eta(\bm{\pi})$, we have
	\begin{align*}
	\eta(\tilde{\bm{\pi}}) - L^2_{\bm{\pi}}(\tilde{\bm{\pi}}) &=  \bm{\rho}_0^T (\tilde{\Pi} - \Pi) (I-\gamma P \tilde{\Pi})^{-1} \bm{A}_{\bm{\pi}} \\
	& = \bm{\rho}_0^T (d \Pi) (I-\gamma P \tilde{\Pi})^{-1} \bm{A}_{\bm{\pi}}
	%& = \bm{\rho}_0^T (d \Pi) \left((I-\gamma P \Pi)^{-1} + (I-\gamma P \Pi )^{-1} (\gamma P d \Pi) (I-\gamma P \Pi )^{-1} \right) \bm{A}_{\bm{\pi}}\\
	%& = \bm{\rho}_0^T (d \Pi) \left(I + (I-\gamma P \Pi )^{-1} (\gamma P d \Pi) \right)(I-\gamma P \Pi )^{-1} \bm{A}_{\bm{\pi}}  \\
	%& =  \textstyle{\bm{\rho}_0^T (d \Pi) \left(I + (\sum_{i=0}^{\infty} \gamma^{i}(P \Pi)^{i} (\gamma P d \Pi)\right) \bm{A}_{\bm{\pi}}}
	%& = \bm{\rho}_0^T (d \Pi) \sum\limits_{i=0}^{\infty} \gamma^{i}(P \Pi)^{i}  (P d \Pi) \bm{A}_{\bm{\pi}}\\
	%& = \bm{\rho}_0^T (d \Pi)(P d \Pi) \bm{A}_{\bm{\pi}} + \bm{\rho}_0^T (d \Pi) \sum\limits_{i=1}^{\infty} \gamma^{i}(P \Pi)^{i}  (P d \Pi) \bm{A}_{\bm{\pi}}\\
	\end{align*}
	Then we have 
	\begin{align*}
	\norm{ \eta(\tilde{\bm{\pi}}) - L^2_{\bm{\pi}}(\tilde{\bm{\pi}}) } &= \norm{\bm{\rho}_0^T (d \Pi) (I-\gamma P \tilde{\Pi})^{-1} \bm{A}_{\bm{\pi}}}\\
	& \leq \norm{\bm{\rho}_0^T} \norm{d \Pi } \norm{\textstyle{\sum_{i=0}^{\infty}} \gamma^{i}(P \tilde{\Pi})^{i}} \norm{\bm{A}_{\bm{\pi}}}\\
	& \leq \frac{\sqrt{2 D_{\mathrm{KL}}^{\max } (\bm{\pi} , \tilde{\bm{\pi}})} }{1-\gamma} \norm{\bm{A}_{\bm{\pi}}}
	%& \leq \norm{\bm{\rho}_0^T}\norm{d \Pi } \norm{I + \textstyle{\sum_{i=0}^{\infty}} \gamma^{i+1}(P \Pi)^{i} P d \Pi} \norm{\bm{A}_{\bm{\pi}}}\\
	%& = \norm{d \Pi} \norm{\bm{A}_{\bm{\pi}}}\norm{d \Pi}/(1-\gamma)\\
	%& \leq \frac{2 D_{\mathrm{KL}}^{\max } (\bm{\pi} , \tilde{\bm{\pi}}) }{1-\gamma} \norm{\bm{A}_{\bm{\pi}}}
	\end{align*}
	
	\subsection{Approximation Function $L^3_{\bm{\pi}}(\tilde{\bm{\pi}})$}
	For $L^3_{\bm{\pi}}(\tilde{\bm{\pi}}) = \eta(\bm{\pi}) + \bm{\rho}_0^T \tilde{\Pi} (I-\gamma P \Pi)^{-1} \bm{A}_{\bm{\pi}}$, it is easy to see that $L^3_{\bm{\pi}}(\bm{\pi}) = 0$.
	%If we set $$L_{\bm{\pi}}(\tilde{\bm{\pi}}) = \eta({\Pi}) + \bm{\rho}_0^T {\Pi} (I-\gamma P \tilde{{\Pi}})^{-1} \bm{A}_{\bm{\pi}} = \eta(\bm{\pi}) + f_{\bm{\pi}}(\tilde{\bm{\pi}})$$
	Moreover, to test \eqref{local function approximation}, we have 
	\begin{equation}
	\begin{aligned}
	\left. d L^3_{\bm{\pi_{\theta_0}}}\left({\bm{\pi}}_{\theta} \right) \right|_{\theta=\theta_{0}} &= \left. d \bm{\rho}_0^T \Pi_{\theta} (I-\gamma P \Pi_{\theta_0})^{-1}  \bm{A}_{\bm{\pi_{\theta_0}}} \right|_{\theta=\theta_{0}} \\
	& = \mathrm{trace} \left( \left. \bm{\rho}_0^T d \Pi_\theta \bm{A}_{\bm{\pi_{\theta_0}}} \right)  \right|_{\theta=\theta_{0}} = \mathrm{trace} \left( \left. \bm{A}_{\bm{\pi_{\theta_0}}} \bm{\rho}_0^T d \Pi_\theta  \right)  \right|_{\theta=\theta_{0}}
	\end{aligned}
	\end{equation}
	And
	\begin{equation}
	\left( \left. \nabla_{\theta} L^3_{\bm{\pi_{\theta_0}}}\left({\bm{\pi}}_{\theta} \right) \right|_{\theta=\theta_{0}} \right)^T = \left. \bm{A}_{\bm{\pi_{\theta_0}}} \bm{\rho}_0^T \frac{\partial \Pi_{\theta}}{\partial \theta}\right|_{\theta=\theta_{0}} \neq \left. \nabla_{\theta} \eta\left({\bm{\pi}}_{\theta} \right) \right|_{\theta=\theta_{0}}
	\end{equation}
	which means $L^3_{\bm{\pi}}(\tilde{\bm{\pi}})$ does not match $\eta(\bm{\pi})$ to the first order. But it is easy to check,
	\begin{equation}
	\left. \nabla_{\theta} L^2_{\bm{\pi_{\theta_0}}}\left({\bm{\pi}}_{\theta} \right) \right|_{\theta=\theta_{0}} + \left. \nabla_{\theta} L^3_{\bm{\pi_{\theta_0}}}\left({\bm{\pi}}_{\theta} \right) \right|_{\theta=\theta_{0}} = \left. \nabla_{\theta} \eta\left({\bm{\pi}}_{\theta} \right) \right|_{\theta=\theta_{0}} = \left. \nabla_{\theta} L^4_{\bm{\pi_{\theta_0}}}\left({\bm{\pi}}_{\theta} \right) \right|_{\theta=\theta_{0}}
	\end{equation}
	which means the gradient of $ L^2_{\bm{\pi_{\theta_0}}}\left({\bm{\pi}}_{\theta} \right)$ and $ L^3_{\bm{\pi_{\theta_0}}}\left({\bm{\pi}}_{\theta} \right)$ is a gradient decomposition of $\eta(\bm{\pi}_\theta)$ at $\bm{\pi}_{\theta_0}$. To calculate the difference between $L^3_{\bm{\pi}}(\tilde{\bm{\pi}})$ and $\eta(\bm{\pi})$, we have
	\begin{align*}
	\eta(\tilde{\bm{\pi}}) - L^3_{\bm{\pi}}(\tilde{\bm{\pi}}) &= \bm{\rho}_0^T \tilde{\Pi} \left((I-\gamma P \tilde{\Pi})^{-1} - (I-\gamma P \Pi)^{-1} \right) \bm{A}_{\bm{\pi}} \\
	& = \bm{\rho}_0^T \tilde{\Pi} \left((I-\gamma P \Pi )^{-1} (\gamma P d \Pi) (I-\gamma P \Pi )^{-1}\right) \bm{A}_{\bm{\pi}}\\
	& = \bm{\rho}_0^T \tilde{\Pi} \left((I-\gamma P \Pi )^{-1} (\gamma P d \Pi) \right) \bm{A}_{\bm{\pi}}
	%& = \bm{\rho}_0^T d \Pi ((I-\gamma P \Pi )^{-1} (\gamma P d \Pi) )\bm{A}_{\bm{\pi}} + \bm{\rho}_0^T ( \sum\limits_{i=0}^{\infty} \gamma^{i}(\Pi P)^{i+1} )d \Pi \bm{A}_{\bm{\pi}} \\
	%& = \bm{\rho}_0^T (d \Pi) \sum\limits_{i=0}^{\infty} \gamma^{i}(P \Pi)^{i}  (P d \Pi) \bm{A}_{\bm{\pi}}\\
	%& = \bm{\rho}_0^T (d \Pi)(P d \Pi) \bm{A}_{\bm{\pi}} + \bm{\rho}_0^T (d \Pi) \sum\limits_{i=1}^{\infty}} \gamma^{i}(P \Pi)^{i}  (P d \Pi) \bm{A}_{\bm{\pi}}\\
	\end{align*}
	Then we have 
	\begin{align*}
	\norm{\eta(\tilde{\bm{\pi}}) - L^3_{\bm{\pi}}(\tilde{\bm{\pi}}) } &= \norm{\bm{\rho}_0^T \tilde{\Pi} \left((I-\gamma P \Pi )^{-1} (\gamma P d \Pi) \right) \bm{A}_{\bm{\pi}}} \\
	& \leq \norm{\bm{\rho}_0^T} \norm{d \Pi} \norm{ \textstyle{\sum_{i=0}^{\infty}} \gamma^{i+1}(P \Pi)^{i} P} \norm{\tilde{\Pi}} \norm{\bm{A}_{\bm{\pi}}} \\
	& = \frac{\gamma \norm{d \Pi} \norm{\bm{A}_{\bm{\pi}}}}{1-\gamma}  \\
	& \leq \frac{\gamma \sqrt{2 D_{KL}(\bm{\pi} , \tilde{\bm{\pi}})}\norm{\bm{A}_{\bm{\pi}}}}{1-\gamma}
	\end{align*}
	\subsection{Approximation Function $L^4_{\bm{\pi}}(\tilde{\bm{\pi}})$}
	
	If we set $$L_{\bm{\pi}}(\tilde{\bm{\pi}}) = \eta(\bm{\pi}) + \bm{\rho}_0^T  (I-\gamma P \Pi)^{-1} \tilde{\Pi} A_{\bm{\pi}} = \eta(\bm{\pi}) + f_{\bm{\pi}}(\tilde{\bm{\pi}})$$
	Then,
	\begin{align*}
	\eta(\tilde{\bm{\pi}}) - L^4_{\bm{\pi}}(\tilde{\bm{\pi}}) &=  \bm{\rho}_0^T  ((I-\gamma P \tilde{\Pi})^{-1} - (I-\gamma P \Pi)^{-1} )\tilde{\Pi} A_{\bm{\pi}}\\
	& = \bm{\rho}_0^T ((I-\gamma P (\Pi + d \Pi) )^{-1} - (I-\gamma P \Pi)^{-1}) (\Pi + d \Pi) A_{\bm{\pi}} \\
	& = \bm{\rho}_0^T (I-\gamma P \Pi )^{-1} (\gamma P d \Pi) (I-\gamma P \Pi )^{-1} d \Pi A_{\bm{\pi}}
	%& = \bm{\rho}_0^T (d \Pi) \textstyle{\sum_{i=0}^{\infty}} \gamma^{i}(P \Pi)^{i}  (P d \Pi) A_{\bm{\pi}}\\
	%& = \bm{\rho}_0^T (d \Pi)(P d \Pi) A_{\bm{\pi}} + \bm{\rho}_0^T (d \Pi) \sum_{i=1}^{\infty} \gamma^{i}(P \Pi)^{i}  (P d \Pi) A_{\bm{\pi}}\\
	\end{align*}
	Then we have 
	\begin{align*}
	\norm{ \eta(\tilde{\bm{\pi}}) - L^4_{\bm{\pi}}(\tilde{\bm{\pi}}) } 
	&= \norm{\bm{\rho}_0^T (I-\gamma P \Pi )^{-1} (\gamma P d \Pi) (I-\gamma P \Pi )^{-1} d \Pi A_{\bm{\pi}}}\\
	& \leq \norm{\bm{\rho}_0^T} \norm{(I-\gamma P \Pi )^{-1}} \norm{\gamma P} \norm{d \Pi } \norm{(I-\gamma P \Pi )^{-1}} \norm{d \Pi} \norm{A_{\bm{\pi}}}\\
	& = \frac{\gamma \norm{d \Pi}^2 \norm{A_{\bm{\pi}}}}{(1-\gamma)^2}\\
	& \leq \frac{2 \gamma D_{KL}(\bm{\pi} , \tilde{\bm{\pi}}) \norm{A_{\bm{\pi}}}}{(1-\gamma)^2}
	\end{align*}
	which is the same as \eqref{inequality KL}.
	
	\section{Other Potential Applications}
	There are some other applications of this set of notations, \eg{}
	\begin{itemize}
		\item If we know $\nabla_{\bm{\pi}_\theta} \eta(\bm{\pi}_\theta)$ and want to get the $\nabla_{\theta} \eta(\bm{\pi}_\theta)$, \ie{} when we know the optimal direction to update policy but we do not know how to control the parameters to make the policy turn to this direction, we can do
		$$\underset{\theta}{\text{min}} \; d(\Pi_{opt}, \Pi_\theta),$$
		where $d(\cdot,\cdot)$ is a distance measure.
		\item Instead of estimating value function $\bm{V_\pi}$, we can estimate the environment dynamics $P$, which is independent of $\bm{\pi}$. Each transition information in the trajectory is valuable no matter the reward signal is detected or not. Then, we can use the estimated $\hat{P}$ to update $\pi$ directly.
	\end{itemize}
	
	\clearpage
	\bibliographystyle{abbrv}
	\bibliography{references}
	
\end{document}